\documentclass{article}

\usepackage[preprint]{neurips_2022}
\usepackage{multirow}

\usepackage[utf8]{inputenc} %
\usepackage[T1]{fontenc}    %
\usepackage{hyperref}       %
\usepackage{url}            %
\usepackage{booktabs}       %
\usepackage{amsfonts}       %
\usepackage{nicefrac}       %
\usepackage{microtype}      %
\usepackage{xcolor}         %
\usepackage{graphicx}
\usepackage{amsmath}
\usepackage[linesnumbered,ruled]{algorithm2e}
\usepackage{algpseudocode}
\usepackage{subcaption}
\usepackage{wrapfig}

\let\oldnl\nl%
\newcommand{\nonl}{\renewcommand{\nl}{\let\nl\oldnl}}%

\usepackage{amsmath,amsfonts,bm}

\def\eqref#1{equation~\ref{#1}}

\def\1{\bm{1}}

\def\vw{{\bm{w}}}
\def\vx{{\bm{x}}}

\def\vz{{\bm{z}}}

\def\mM{{\bm{M}}}

\def\mW{{\bm{W}}}
\def\mX{{\bm{X}}}
\def\mY{{\bm{Y}}}
\def\mZ{{\bm{Z}}}

\DeclareMathAlphabet{\mathsfit}{\encodingdefault}{\sfdefault}{m}{sl}
\SetMathAlphabet{\mathsfit}{bold}{\encodingdefault}{\sfdefault}{bx}{n}

\title{Continuous-time Particle Filtering for Latent Stochastic Differential Equations}

\author{%
  Ruizhi Deng$^{1, 2}$\thanks{This work was done during an internship at Borealis AI. Correspondance to wsdmdeng@gmail.com.} ~~~~Greg Mori$^{1, 2}$~~~~Andreas M. Lehrmann$^{1}$\\
  $^1$Borealis AI~~~~$^2$Simon Fraser University\\
}

\begin{document}

\maketitle

\begin{abstract}
  Particle filtering is a standard Monte-Carlo approach for a wide range of sequential inference tasks. The key component of a particle filter is a set of particles with importance weights that serve as a proxy of the true posterior distribution of some stochastic process. In this work, we propose continuous latent particle filters, an approach that extends particle filtering to the continuous-time domain. We demonstrate how continuous latent particle filters can be used as a generic plug-in replacement for inference techniques relying on a learned variational posterior. Our experiments with different model families based on latent neural stochastic differential equations demonstrate superior performance of continuous-time particle filtering in inference tasks like likelihood estimation and sequential prediction for a variety of stochastic processes.
\end{abstract}

\section{Introduction}
Over the last years neural architectures based on latent stochastic differential equations (latent SDEs;~\cite{tzen2019neural, li2020scalable}) have emerged as expressive models of continuous-time dynamics with instantaneous noise. As a temporal backbone in continuously-indexed normalizing flows~\cite{deng2021continuous} they have also proven to be a powerful latent representation of non-Markovian dynamics. Further extensions to high-dimensional time-series~\cite{hasan2020identifying} and infinitely-deep Bayesian networks~\cite{xu22} have demonstrated applications to biology and computer vision.

In the latent SDE framework observations can be viewed as partial realizations of an intrinsically continuous process. While the resulting architecture is very flexible, direct computation of the marginal observation likelihood is typically not tractable and approximate techniques are required for training and inference. A common approach in the literature is a variational approximation: \cite{li2020scalable} introduce an approximate posterior SDE and formulate an ELBO objective~\cite{kingma2013auto} for SDEs. \cite{deng2021continuous} build upon this framework but decompose the marginal observation likelihood into intervals defined by the observation times to enable non-Markovian dynamics and efficient online inference. In both cases, an importance-weighted autoencoder (IWAE;~\cite{burda2015importance}) can be used to obtain a tighter variational bound.

In this paper, we approach inference in latent SDEs from a different perspective and adapt traditional particle filtering to the continuous-time domain of latent SDEs. First, we introduce a rigorous mathematical framework that defines the structure of particles, weights, and updates in the context of latent SDEs. Then, we demonstrate how important inference tasks, such as likelihood estimation and sequential prediction, can be cast in the form of expectations over posterior distributions that are amenable to particle filtering. The benefits of this approach are two-fold: (1) during likelihood estimation, the resampling step of the particle filter drops samples with smaller weights and keeps samples with larger weights, leading to \emph{higher sample efficiency} than IWAE estimation; (2) during sequential prediction, the particle filter converges in expectation to the true posterior, leading to \emph{higher prediction accuracy} than a potentially restrictive learned approximation. The proposed method serves as a direct plug-in replacement for IWAE-based variational inference in latent SDEs. Our experiments validate continuous-time particle filtering for latent SDEs on stochastic processes with a broad range of properties, including geometric Brownian motion and SDEs with linear, multi-dimensional, coupled, and non-Markovian dynamics.

\paragraph{Contributions.} In summary, we make the following contributions: (1) we propose a mathematically rigorous extension of traditional particle filtering to the continuous-time domain; (2) we demonstrate the use of the proposed continuous-time particle filtering framework as a plug-in replacement for importance-weighted variational inference in latent SDEs; (3) we evaluate the resulting estimator on likelihood estimation and sequential prediction tasks, and demonstrate superior sample-efficiency and accuracy on a broad set of stochastic processes.

\section{Preliminaries: Latent SDEs for Time-Series Modeling}
As our continuous-time particle filtering approach is an inference algorithm generally applicable to models based on the existing latent SDE framework~\cite{li2020scalable, deng2021continuous}, we will first provide an abstract description of the main ideas behind these models using a unified notation.
Let $\{(t_i, \vx_{t_i})\}_{i=1}^n$ denote a sequence of observations on the time grid $t_1<t_2<\dots<t_n$, with $t_i \in (0, T)$, and $\vx_{t_i}$ be an $m$-dimensional observation at time point $t_i$. 
Furthermore, let $(\Omega, \mathcal{F}_t, P)$ be a filtered probability space on which $\mW_t$ is a $d$-dimensional Wiener process. A latent stochastic differential equation modeling this observation sequence consists of two stochastic processes: a $d$-dimensional latent process $\mZ_t$ defined by a stochastic differential equation
\begin{equation}
    \textrm{d}\mZ_t = \mu_\theta(\mZ_t, t)\ \textrm{d}t + \sigma_\theta(\mZ_t, t)\ \textrm{d}\mW_t,
    \label{eq:latent_process}
\end{equation}
and an $m$-dimensional observable process $\mX_t=f_\theta(\mZ_t)$ obtained by decoding the latent process with a decoder $f$, where $\theta$ denotes all parameters of the model. It is worth noting that each stochastic differential equation can be viewed as a mapping of Wiener process paths to SDE paths, with the Wiener process being the actual source of stochasticity.
To sample observation sequences $\{\vx_{t_i}\}_{i=1}^n$ from the model, we first obtain samples $\{\vz_{t_i}\}_{i=1}^n$ from the joint distribution of $\{\mZ_{t_i}\}_{i=1}^n$ induced by the process $\mZ_t$ on the given time grid by solving Eq.(\ref{eq:latent_process}). The sample sequence $\{\vz_{t_i}\}_{i=1}^n$ is then decoded into a joint observational distribution over $\{\mX_{t_i}\}_{i=1}^n$ conditioned on the values of $\{\vz_{t_i}\}_{i=1}^n$. For some models (e.g.,~\cite{li2020scalable}), the sample sequence $\{\vz_{t_i}\}_{i=1}^n$ can be decoded independently at each time step $t_i$,
\begin{equation}
    p_{\mX_{t_1},\dots,\mX_{t_n}|\vz_{t_1},\dots,\vz_{t_n}}(\vx_{t_1}, \vx_{t_2},\dots,\vx_{t_n}|\vz_{t_1}, \vz_{t_2},\dots,\vz_{t_n}) = \prod_{i=1}^n p_{\mX_{t_i}|\vz_{t_i}}(\vx_{t_i}|\vz_{t_i}),
\end{equation}
while other models (e.g.,~\cite{deng2021continuous}) assume more complex dependency structures between the $\mZ_{t_i}\!$'s and $\mX_{t_i}\!$'s.

\subsection{Importance Weighting for Latent SDEs} Similar to other latent variable models~\cite{kingma2013auto}, computing the marginal observation likelihood in latent SDEs is intractable and relies on a learned variational posterior process for training and inference tasks like likelihood estimation and sequential prediction. By optimizing the evidence lower bound (ELBO) of the log-likelihood, the variational posterior process is explicitly encouraged to reconstruct the observations from their latent representations with high probability. The learned posterior process can therefore be used for a variety of downstream inference tasks, including likelihood estimation and sequential prediction. Learning the variational posterior process and running inference in latent SDE models require computing importance weights of latent trajectories. Likewise, importance weighting is also an essential component of particle filtering. In the remainder of this section we derive the importance weight induced by a variational posterior process and lay the foundation for the particle filtering approach discussed in Section~\ref{sec:ctpf}.

Given the observations $\{(t_i, \vx_{t_i})\}_{i=1}^n$, the posterior process $\tilde{\mZ}_t$ in a latent SDE model is characterized by a stochastic differential equation with an observation-dependent drift term $\mu_\phi$ and a shared variance term $\sigma_\theta$,
\begin{equation}
    \textrm{d}\tilde{\mZ}_t = \mu_\phi(\tilde{\mZ}_t, t)\ \textrm{d}t + \sigma_\theta(\tilde{\mZ}_t, t)\ \textrm{d}\mW_t,
    \label{eq:post_process}
\end{equation} 
such that Eq.(\ref{eq:post_process}) satisfies Novikov's condition
\begin{equation}
    \mathbb{E}\left[\exp(\int_0^T \frac{1}{2} \left|u(\tilde{\mZ_t}, t)\right|^2\ \textrm{d}t)\right] < \infty,
\end{equation}
with $\sigma(z, t)u(z, t) = \mu_\phi(z, t) - \mu_\theta(z, t)$. The parameters of the drift function $\phi$ are produced by the observations $\{(t_i, \vx_{t_i})\}_{i=1}^n$ so the posterior process can be conditioned on the observations. The parameters of the variance function $\theta$ are the same as in the prior latent process (Eq.~\ref{eq:latent_process}) to allow computation of the importance weight between the prior and posterior process:
By Girsanov's Theorem~\cite[Theorem 8.6.4]{oksendal2013stochastic}, we can equip the measurable space $(\Omega, \mathcal{F}_t)$ with a probability distribution $Q$, such that $\tilde{\mW}_t = \int_0^t u(\tilde{\mZ_t}, t)\ \textrm{d}t + \mW_t$ is another Wiener process, with $\mW_t$ defined on $(\Omega, \mathcal{F}_t, Q)$. Moreover, computing the importance weight (Radon-Nikodym derivative) between the distributions $Q$ and $P$ is tractable and we denote the importance weight for a trajectory of length $t$ by $\mM_t(\omega)$. Specifically, given a sample $\omega\in \Omega$, we have
\begin{equation}
    \mM_t(\vz_t) = \exp(-\int_0^t \frac{1}{2} |u(\vz_s, s)|^2\ \textrm{d}s - \int_0^t u(\vz_s, s)^T\ \textrm{d}\mW_s(\omega)).
    \label{eq:importance}
\end{equation}
Since $\tilde{\mW}_t$ is also a Wiener process, we can replace $\mW_t$ in the original prior process (Eq.(\ref{eq:latent_process})) with $\tilde{\mW}_t$ without changing the law (push-forward measure) of $\mZ_t$,
\begin{equation}
    \textrm{d}\mZ_t = \mu_\theta(\mZ_t, t)\ \textrm{d}t + \sigma_\theta(\mZ_t, t)\ \textrm{d}\tilde\mW_t = \mu_\phi(\mZ_t, t)\textrm{d}t + \sigma_\theta(\mZ_t, t)\textrm{d}\mW_t,
\end{equation}
for $\mW_t$ defined on $(\Omega, \mathcal{F}_t, Q)$.

Sampling directly from the unknown distribution $Q$ is challenging. However, to obtain a Monte-Carlo estimate of an expectation with respect to $Q$, we can first sample $\mW_t(\omega)$ using the distribution $P$ and then apply the importance weight $\mM_t$ defined by Eq.(\ref{eq:importance}), leading to
\begin{equation}
    \mathbb{E}_P[f(\{\mZ_{t_i}\}_{i=1}^n)]
    =\mathbb{E}_Q[f(\{\tilde\mZ_{t_i}\}_{i=1}^n)]
    =\mathbb{E}_P[f(\{\tilde\mZ_{t_i}\}_{i=1}^n)\mM_t].
    \label{eq:exp}
\end{equation}
The posterior process $\tilde{\mZ}_t$ and the importance weighting process $\mM_t$ can be concatenated and together form the solution to an augmented version of the SDE characterizing the posterior process (Eq.~\ref{eq:post_process}).

\section{Continuous-Time Particle Filtering}\label{sec:ctpf}
With the background knowledge on latent SDEs and importance weighting introduced, we will now formally present our continuous-time particle filtering approach (Section~\ref{sec:ct_particle_filtering}) along with two applications to inference in latent SDEs (Section~\ref{sec:inference_likelihood}).
\begin{figure}
    \centering
    \includegraphics[width=\textwidth]{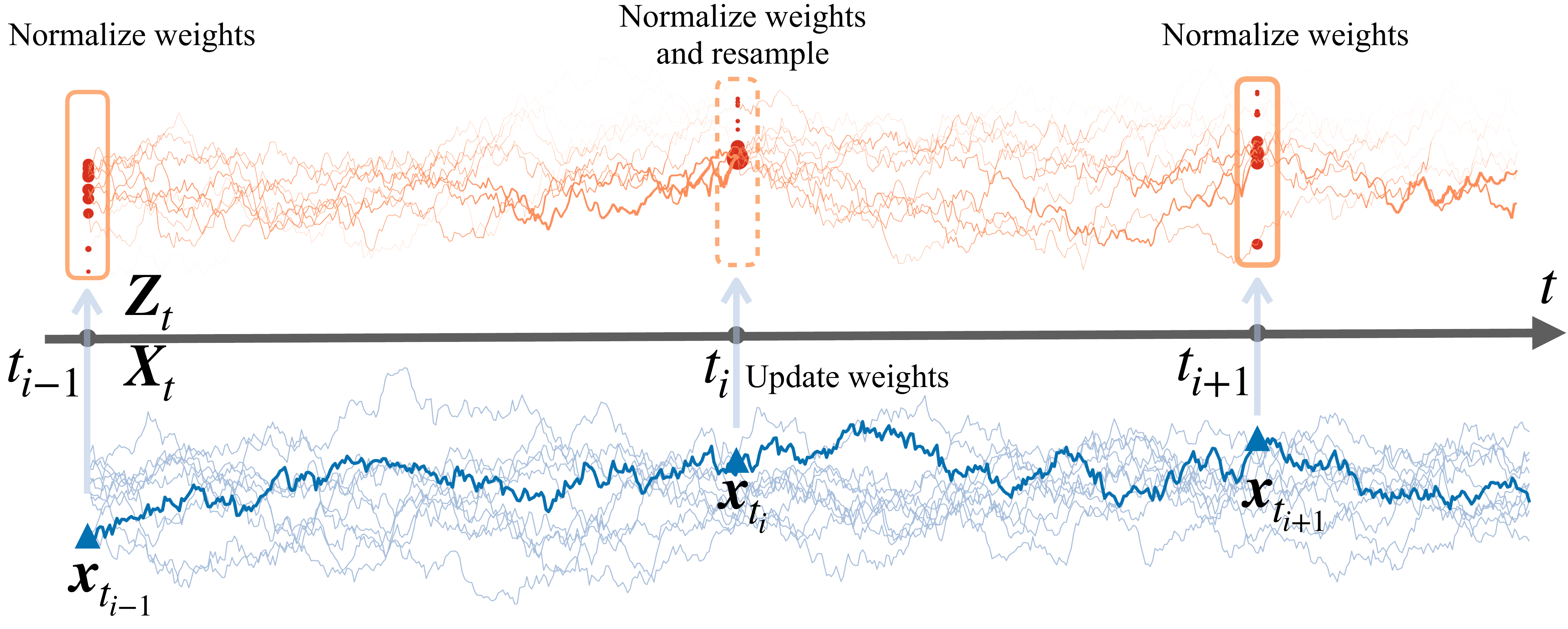}
    \caption{
    {\bf Overview of Continuous-Time Particle Filtering.} Given discrete observations (blue triangles) of a sample trajectory of the observed process $\mX_t$, the importance weights of latent trajectories (orange trajectories), represented by line thickness, are first updated by observation likelihood (blue arrows) and normalized (orange boxes). If the normalized weights (orange circles) are concentrated on a small subset of the particles, the particles will be resampled (dashed orange box) with the normalized weights set to be uniform in value. The normalized weights are used as initial weights of the latent trajectories in the following time interval.
    }
    \label{fig:fig1}
\end{figure}

\subsection{Particles and Weights for Continuous-time Latent SDEs}\label{sec:ct_particle_filtering}
The core intuition behind particle filtering is to use a set of particles (i.e., samples from a distribution) as a proxy of the posterior distribution of a stochastic process given sequential observations. Three important questions need to be answered when extending particle filtering to the latent SDE framework: (1) what are the particles in a continuous-time particle filter; (2) what are the prior and importance weighting distributions; and (3) how should the particle weights be updated.

{\bf Continuous-time Particles.} In the latent process of latent SDE models, the expectation in Eq.(\ref{eq:exp}) is taken over the distribution of a Wiener process. It is thus natural to use sample trajectories of a Wiener process as particles in the particle filter. Even though Wiener processes are usually defined on probability spaces with continuous filtration, the resampling step of a particle filter is a discrete event. As a consequence, we need to give the Wiener process and the continuous-time stochastic processes induced by latent SDEs sequential structure to define particles in a continuous-time latent SDE setting. To this end, we leverage the piece-wise construction of Wiener processes proposed in~\cite{deng2021continuous}: given the time grid $0 = t_0<t_1<\dots<t_n=T$, where each $t_i$ is the observation time point of $\vx_{t_i}$, we can obtain a sample trajectory of a Wiener process $\mW_t$ of length $T$ by sampling trajectories from $n$ independent Wiener processes $\mW_t^{(i)}$, each with length $t_i - t_{i-1}$, defined on the probability space $(\Omega^{(i)}, \mathcal{F}^{(i)}_t, P^{(i)})$ and add them together,
\begin{equation}
    \mW_t(\omega^{(1)}, \omega^{(2)}, \dots, \omega^{(n)}) = \sum_{\{i: t_i<t\}}\mW_{t_i- t_{i-1}}^{(i)}(\omega^{(i)}) + \mW_{t- t_{i^*}}^{(i^*)}(\omega^{(i^*)}),
\end{equation} where $i^* = \max\{i: t_i<t\} + 1$ and $\omega^{(i)}\in \Omega^{(i)}$.
We can solve Eq.(\ref{eq:latent_process}) in a similar piece-wise manner,
\begin{equation}
    \mZ_t =  \sum_{\{i: t_i<t\}} \mZ_{t_i}+ \int_{t_{i^*}}^{t} \mu_\theta(\mZ_s,s )\ \textrm{d}s + \int_{t_{i^*}}^{t} \sigma_\theta(\mZ_s,s )\ \textrm{d}\mW^{(i^*)}_{s - t_{i^*}},
\end{equation} and obtain the posterior process $\tilde{\mZ}_t$ (Eq.(\ref{eq:post_process})) by defining $(\Omega^{(i)}, \mathcal{F}^{(i)}_t, Q^{(i)})$ and posterior process parameters $\phi^{(i)}$. Moreover, the distributions $Q^{(i)}$ and parameters $\phi^{(i)}$ can be conditioned on samples of $\{\tilde{\mZ}_{t_j}\}$ for $j\leq i$ without changing the law of $\tilde{\mZ}_t$.
We refer to the appendix for additional details on the piecewise construction of the posterior process $\tilde{\mZ}_t$. In summary, each particle in the continuous-time particle filter will be represented as a sequence of $\omega^{(i)}$'s, where each $\omega^{(i)}$ is a sample from $\Omega^{(i)}$.

{\bf Prior and Importance Weighting.} The importance weight distribution in a particle filter, also called the proposal distribution, should be easy to sample from. Therefore, we use $(\Omega^{(i)}, \mathcal{F}_t^{(i)}, P^{(i)})$ as importance weight distribution to sample $\omega^{(i)}$ for the interval $[t_{i-1}, t_i]$. Note that this choice of importance weight distribution admits a sequential structure in the sense that the distributions before $t_i$ will not be modified given future observations after $t_i$. Moreover, the evidence lower bound also encourages the variational posterior process to reconstruct observations with high likelihood from the Wiener process.
We make the choice of $Q^{(i)}$ conditioned on samples from $\{\tilde{\mZ}_{t_j}\}_{j<i}$ as the prior distribution. 
However, it is worth noting that, despite being observation dependent, the $Q^{(i)}$'s induce the same finite-dimensional (prior and posterior) distributions of $\tilde{\mZ}_t$ as the finite-dimensional distribution of $\mZ_t$ induced by the $P^{(i)}$'s. As a result, we can rewrite Eq.(\ref{eq:exp}) using this piece-wise approach as follows: 
\begin{align}
\begin{split}
      &\mathbb{E}_{P^{(1)}\times\dots\times P^{(i)}\times\dots\times P^{(n)}}\left[f\left(\left\{\mZ_{t_i}\right\}_{i=1}^n\right)\right] \\[4pt]
   =\ &\mathbb{E}_{P^{(1)}}\left[\dots\mathbb{E}_{P^{(i)}}\left[\dots \mathbb{E}_{P^{(n)}}\left[f\left(\left\{\mZ_{t_i}\right\}_{i=1}^n\right)\right]\dots\right]\dots\right] \\[4pt]
    =\ &\mathbb{E}_{Q^{(1)}|\{\tilde{\mZ}_{t_1}\}}\left[\dots\mathbb{E}_{Q^{(i)}|\{\tilde\mZ_{t_k}\}_{k=1}^{i-1}}\left[\dots \mathbb{E}_{Q^{(n)}|\{\tilde\mZ_{t_k}\}_{k=1}^{n-1}}\left[f(\{\tilde\mZ_{t_i}\}_{i=1}^n)\right]\dots\right]\dots\right]\\[4pt]
    =\ &\mathbb{E}_{P^{(1)}}\left[\dots\mathbb{E}_{P^{(i)}}\left[\dots \mathbb{E}_{P^{(n)}}\left[f(\{\tilde\mZ_{t_i}\}_{i=1}^n)\mM^{(n)}\right]\dots \mM^{(i)}\right]\dots\mM^{(1)}\right],
\end{split}
\label{eq:exp_piece}
\end{align}
where $P^{(1)}\times\dots\times P^{(i)}\times\dots\times P^{(n)}$ is the product distribution of all $P^{(i)}$'s from which we sample the $\mW_t^{(i)}$'s and $\mM^{(i)}$ is shorthand for $\mM^{(i)}_{t_i - t_{i-1}}$, i.e., the $i$-th importance weight term between distributions $P^{(i)}$ and $Q^{(i)}|\{\tilde\mZ_{t_k}\}_{k=1}^{i-1}$ defined by Eq.(\ref{eq:importance}), or equivalently $\frac{\textrm{d} Q^{(i)}|\{\tilde\mZ_{t_k}\}_{k=1}^{i-1}}{\textrm{d}P^{(i)}}$.

{\bf Particle Updates.} With the prior and importance weight distribution specified, we can now present how samples are obtained and weights are updated in continuous-time particle filters given a sequence of observations $\vx_{t_i}$.
Let $N$ be the number of particles in a particle filter and $\{(\{\omega^{(k)}_j\}_{k=1}^{i},\vw_j^{(i)})\}_{j=1}^N$ denote the set of particles after the $i$-th observation $\vx_{t_i}$ and update, where $\omega^{(k)}_j$ is a sample from $(\Omega^{(k)}, \mathcal{F}_t^{(k)}, P^{(k)})$ for each $j$ and $\vw_j^{(i)}$ is the weight of the $j$-th particle up to time $t_i$. Following the standard formulation of particle filtering, we initialize the set of particles as $\{(\{\}, \vw^{(0)}_j)\}_{j=1}^N$, i.e., each particle has no sample, denoted as $\{\}$, and an initial weight of $\vw^{(0)}_j:=\frac{1}{N}$. 
The set of particles with their weights are updated by every new observation.
Given the set of particles $\{(\{\omega^{(k)}_j\}_{k=1}^{i-1},\vw_j^{(i-1)})\}_{j=1}^N$ at time point $t_{i-1}$ and the $i$-th observation $\vx_{t_i}$, the particles are updated as follows:
\begin{align}
\begin{split}
    \{\omega^{(k)}_j\}_{k=1}^{i}&\leftarrow   \textit{CONCAT}\left(\{\omega^{(k)}_j\}_{k=1}^{i-1}, \{\omega_j^{(i)}\}\right), \\[4pt]
    \tilde{\vw}_j^{(i)}&\leftarrow \vw_j^{(i-1)} p(\vx_{t_i}|\{\vx_{t_k}\}_{k=1}^{i-1}, \{\omega^{(k)}_j\}_{k=1}^{i})\mM_j^{(i)}(\omega^{(i)}_j|\{\omega^{(k)}_j\}_{k=1}^{i-1}), \\[4pt]
    \vw_j^{(i)}&\leftarrow  \tilde{\vw}_j^{(i)}/\ \textstyle{\sum_{j=1}^N} \tilde{\vw}_j^{(i)}
\end{split}
\end{align}
where $\mM_j^{(i)}$ is the importance weight for sample $\omega_j^{(i)}$ in the interval ${t_i - t_{i-1}}$ and $p(\vx_{t_i}|\{\omega^{(k)}_j\}_{k=1}^{i},\{\vx_{t_k}\}_{k=1}^{i-1})$ is the observation likelihood used to update the weights of particles conditioned on previous observations and latent samples. The likelihood term can usually be written as $p(\vx_{t_i}|\{\vx_{t_k}\}_{k=1}^{i-1}, \{\vz_{t_k, j}\}_{k=1}^{i})$, with $\vz_{t_k, j}$ defined by $\omega^{(k)}_j$ through the latent SDE.
When certain criteria are met~\cite{doucet2009tutorial}, we resample the particles from the categorical distribution defined by the particle weights and all the weights $\vw_j^{(i)}$ are reset to $\frac{1}{N}$. The process of sampling and updating particle weights is summarized in Algorithm~\ref{algo:CTPF} and visualized in Figure~\ref{fig:fig1}. 

\begin{algorithm}[t]
\SetAlgoLined
\SetKwInput{Input}{input}
\SetKwInput{Output}{output}
\SetKwInput{Init}{Initialization}
\SetKwInput{Blank}{}
\SetKwInput{Ret}{Return}

\tabcolsep=0pt
\nonl \begin{tabular}{@{}ll}
    \Input{}&Observation sequence with time points $\{(t_i, \vx_{t_i})\}_{i=1}^n$; number of particles $N$ \;\\
    & Latent process drift function with adaptable parameters $\mu(z, t; \text{parameters}):\mathbb{R}^M\times \mathbb{R}\rightarrow \mathbb{R}^M$\;\\
    & Latent process variance function $\sigma(z, t):\mathbb{R}^M\times \mathbb{R}\rightarrow \mathbb{R}^M\times \mathbb{R}^M$\;\\
    & Parameters for the drift function $\mu$ in the original prior process of latent SDE $\theta$\;\\
    & A function generating parameters for drift function from sequence parameters \textit{PARAM\_GEN}$(\{\vz_{t_k}\}_{k=1}^i)$\;\\
    & A sampler of an $M$-dimesional Wiener process trajectory, with time length $t$ as input \textit{SAMPLER}$(t)$\;\\
    & Conditional likelihood evaluation function of observations $p(\vx_{t_i}|\{\vz_{t_k}\}_{k=1}^i, \{\vx_{t_k}\}_{k=1}^{i-1})$\;\\
    & Initial state of the latent process $\vz_{t_0}$\;\\
    & A boolean resampling condition function \textit{RESAMPLE\_CON}$(\{\vw_j\}_{j=1}^N)$\;\\
    \Output{}& A set of particles with weights $\{(\{\omega^{(k)}_j\}_{k=1}^{n},\vw_j^{(n)})\}_{j=1}^N$\;\\
\end{tabular}
\BlankLine
 $\mathbf{Initialization : }$ $\vw^{(0)}_j\leftarrow \frac{1}{N}, \vz_{t_0,j}=\vz_{t_0} ~\textbf{for } j\leftarrow 1 \textbf{ to } N$; particle\_set$\ \leftarrow\{(\{\}, \vw^{(0)}_j)\}_{j=1}^N$ \;
 
 \For{$i\leftarrow 1$ \KwTo $n$}{
 \For{$j\leftarrow 1$ \KwTo $N$}{
 $\omega^{(i)}_j\leftarrow$ \textit{SAMPLER}$(t_i - t_{i-1})$\;
 $\phi^{(i)}_j\leftarrow$ \textit{PARAM\_GEN}$(\{\vz_{t_i, j}\}_{k=0}^{i-1})$\;
 \tcc*[h]{Solve the stochastic differential equation in the interval $[t_{i-1}, t_i]$ given the Wiener process path sample $\omega^{(i)}_j$.}\\
 $\mM^{(i)}_j, \vz_{t_i,j}\leftarrow$ \textit{AUG\_SDESOLVE}$(t_{i-1}, t_i, \vz_{t_{i-1}}, \omega^{(i)}_j, \sigma(\cdot, \cdot), \mu(\cdot, \cdot, \theta), \mu(\cdot, \cdot, \phi^{(i)}_j))$\;
 \tcc*[h]{\textit{AUG\_SDESOLVE} not only solves the SDE but also computes the importance weights of sample between the prior and proposal distributions. }\\
 $\{\omega^{(k)}_j\}_{k=1}^{i} \leftarrow$ \textit{CONCAT}$(\{\omega^{(k)}_j\}_{k=1}^{i-1}, \{\omega^{(i)}_j\})$\;
 $\tilde{\vw}_j^{(i)} \leftarrow \vw_j^{(i)}p(\vx_{t_i}|\{\vz_{t_k, j}\}_{k=1}^i, \{\vx_{t_k}\}_{k=1}^{i-1})\mM_j^{(i)}$\;
 }
 \tcc*[h]{Normalize the weights.}\\

$\vw_j^{(i)}\leftarrow  \frac{\tilde{\vw}_j^{(i)}}{\sum_{j=1}^N \tilde{\vw}_j^{(i)}}\ \textbf{for}\ j \leftarrow 1\ \textbf{to}\ N$\;

\If{RESAMPLE\_CON$(\{\omega^{(i)}_j\}_{j=1}^N)$}{
\tcc*[h]{Resample the particles from categorical distributions defined by the importance weights and reset the weights.}
$\{\{\omega^{(k)}_j\}_{k=1}^{i}\}_{j=1}^N\leftarrow CAT\_SAMPLE(\{\{\omega^{(k)}_j\}_{k=1}^{i}\}_{j=1}^N,\{\vw_j^{(i)}\}_{j=1}^N )$\;
$\vw^{(i)}_j\leftarrow \frac{1}{N}~\textbf{for}~j\leftarrow 1~\textbf{to}~N$\;
}
particle\_set $\leftarrow \{(\{\omega^{(k)}_j\}_{k=1}^{i},\vw_j^{(i)})\}_{j=1}^N$\;
}

\Ret{particle\_set}
 \caption{Continuous-Time Particle Filter}
 \label{algo:CTPF}
\end{algorithm}

\subsection{Continuous-time Particle Filter for Inference Tasks}
Many inference tasks relying on the posterior distribution can be expressed as an expectation w.r.t.~certain functions over the posterior distribution of $Q$ conditioned on observations $\{\vx_{t_i}\}_{i=1}^n$, i.e., $\mathbb{E}_{Q|\{\vx_{t_i}\}_{i=1}^n}[f(\{\tilde{\mZ}_{t_i}\}_{i=1}^n)]$. As the set of particles with their weights is a proxy of the posterior distribution $Q|\{\vx_{t_i}\}_{i=1}^n$, the weighted average of the function over the particles is a Monte-Carlo integration and an estimator of the expectation. We present two applications of continuous-time particle filtering based on this principle: likelihood estimation and sequential prediction.

\paragraph{Continuous-time Particle Filtering for Likelihood Estimation.}\label{sec:inference_likelihood}
In latent variable models it is common practice to approximate the observation log-likelihood using an IWAE bound with multiple latent samples~\cite{burda2015importance}. When applied to latent SDE models~\cite{li2020scalable,deng2021continuous}, the IWAE bound can be viewed as a specific instance of sequential importance sampling. One concern with sequential importance sampling is decreasing sampling efficiency as time increases, e.g., as a result of importance weights becoming skewed over time, with most weights concentrated on a few samples. The resampling step of particle filters can remove samples with smaller importance weights while preserving the ones with larger importance weights. Given a sequence of observations $\{\vx_{t_i}\}_{i=1}^n$, we take the integrand function $f$ at time $t_{i-1}$ to be
\begin{equation}
    p(\vx_{t_i}|\{\vx_{t_k}\}_{k=1}^{i-1}, \{\mZ_{t_k}\}_{k=1}^{i-1})=\mathbb{E}_{Q^{(i)}|\{\tilde\mZ_{t_k}\}_{k=1}^{i-1}}[p(\vx_{t_i}|\{\vx_{t_k}\}_{k=1}^{i-1}, \{\tilde{\mZ}_{t_k}\}_{k=1}^{i})\mM_{t_i - t_{i-1}}^{(i)}],
    \label{eq:likelihood}
\end{equation} where each $\tilde{\mZ}_{t_k}$ is a function of $\{\omega^{(l)}\}_{l=1}^k$. The samples used for estimating the expectation in Eq.(\ref{eq:likelihood}) can in turn be reused to update the particles at time $t_i$.

\paragraph{Continuous-time Particle Filtering for Sequential Prediction.}\label{sec:inference_prediction}
Many sequential latent variable models~\cite{rubanova2019latent, li2020scalable, deng2020modeling} rely solely on the proposal distribution for forecasting 
and completely discard the prior distribution, which also defines the true posterior distribution. Using particle filtering for forecasting not only reintroduces the prior distribution to the inference task but also makes use of a proxy version of the true posterior instead of an arbitrary proposal distribution. In particular, we are interested in the sequential prediction task of estimating the expectation of $\mX_{t_{i+1}}$ conditioned on the observations $\{\vx_{t_k}\}_{k=1}^i$. It can be formulated as 
\begin{equation}
    \mathbb{E}_{Q|\{\vx_{t_k}\}_{k=1}^i}\left[
    \mathbb{E}_{\mX_{t_{i+1}}|\tilde{\mZ}_{t_i}, \{\vx_{t_k}\}_{k=1}^i}[\vx_{t_{i+1}}]
    \right],
\end{equation}
where the expected value of $\mX_{t_{i+1}}$ conditioned on previous observations and the value of $\tilde{\mZ}_{t_i}$ is the integrand of interest. This expectation can, in turn, be estimated by sampling $\mZ_{t_i}$ from the prior process and averaging the expectation of $\mX_{t_{i+1}}$ conditioned on $\mZ_{t_i}$ and $\{\vx_{t_k}\}_{k=1}^i$.

\section{Experiments}
To demonstrate the general applicability and effectiveness of our proposed particle filtering approach, we apply it to likelihood estimation and sequential prediction tasks in latent SDE~\cite{li2020scalable} and CLPF~\cite{deng2021continuous} models. We compare the performance of these models against the exact same models without particle filtering. For the baseline models, we use an IWAE approximation~\cite{burda2015importance} of the observation likelihood for likelihood estimation and the expectation over the variational posterior process conditioned on observations for sequential prediction.

\subsection{Datasets}

We train and evaluate all models using asynchronous sequential samples simulated from four common continuous-time stochastic processes, i.e., the observation time stamps are on an irregular time grid; please refer to the Appendix for additional parameter settings.

{\bf Geometric Brownian Motion (GBM).} Geometric Brownian motion is a stochastic process satisfying $\textrm{d}\mX_t = \mu\mX_t\ \textrm{d}t + \sigma\mX_t\ \textrm{d}\mW_t$, i.e., the logarithm of geometric Brownian motion is Brownian motion. %

{\bf Linear SDE (LSDE).} 
In a linear SDE the drift term is a linear transformation and the variance term is a deterministic function of time $t$; it can be characterized as $\textrm{d}\mX_t = (a(t)\mX_t + b(t))\ \textrm{d}t + \sigma(t)\ \textrm{d}\mW_t$.%

{\bf Continuous AR(4) Process (CAR).} The continuous autoregressive process of $4$-th order can be viewed as the 1-dim. projection of the following 4-dim. stochastic differential equation:
   \begin{align}
        \begin{split}
            \mX_t &= [d, 0, 0, 0]\mY_t, \\
             \textrm{d}\mY_t &= A\mY_t\ \textrm{d}t + e \ \textrm{d}\mW_t,
        \end{split}, \quad \mbox{ where } A = \begin{pmatrix} {\bf 0} & I_3 \\
        a_1 & \begin{matrix} a_2 & a_3 & a_4\end{matrix}
        \end{pmatrix}.
    \end{align}

{\bf Stochastic Lorenz Curve (SLC).} The stochastic Lorenz curve is a three-dimensional continuous-time stochastic process characterized by the following set of stochastic equations:
    \begin{align}
        \begin{split}
            \textrm{d}\mX_t &= \sigma(\mY_t - \mX_t)\ \textrm{d}t + \alpha_x\ \textrm{d}\mW_t, \\
            \textrm{d}\mY_t &= (\mX_t(\rho - \mZ_t) - \mY_t)\ \textrm{d}t + \alpha_y\ \textrm{d}\mW_t, \\
            \textrm{d}\mZ_t &= (\mX_t\mY_t - \beta\mZ_t)\ \textrm{d}t + \alpha_z\ \textrm{d}\mW_t. \\
        \end{split}
    \end{align}

\subsection{Experiment Setup}
The observation time points of the asynchrounous sequences are sampled from a homogeneous Poisson process with intensity $\lambda$. Please see Section~\ref{sec:additional_experiment_details} in the appendix for more details. After training, the weights of the models are fixed and remain the same whether the particle filter is used during inference or not. Following the experiment settings of CLPF~\cite{deng2021continuous}, we train the models using asynchronous sequences with sparse observations ($\lambda = 2$ for GBM, LSDE, and CAR; $\lambda = 20$ for SLC) but use both sparse and dense sequences during evaluation by controlling the intensity values of the test observation process ($\lambda = [2,20]$ for GBM, LSDE, and CAR; $\lambda = [20,40]$ for SLC). We refer to the supplementary material for additional information about our experiment settings.%

\begin{table}[t]
    \caption{\small {\bf Likelihood Evaluation}. We report and compare negative log-likelihoods (NLLs) of observations estimated using an IWAE approximation and our proposed particle filter. We evaluate both techniques on four stochastic processes with two different observation intensity values $\lambda$. [GBM: geometric Brownian motion (ground truth NLLs: $[\lambda=2,\lambda=20]= [0.388, -0.788]$); LSDE: linear SDE; CAR: continuous auto-regressive process; SLC: stochastic Lorenz curve]}   
    \label{tab:synthetic}
    \centering
    \small
    \scalebox{0.85}{
    \begin{tabular}{l@{\extracolsep{3pt}}rrrrrrrr@{}}
        \toprule
         \multirow{2}{*}{Model}
         &\multicolumn{2}{c}{GBM}&\multicolumn{2}{c}{LSDE}&\multicolumn{2}{c}{CAR}&\multicolumn{2}{c}{SLC}\\
         \cmidrule{2-3}
         \cmidrule{4-5}
         \cmidrule{6-7}
         \cmidrule{8-9}
         &$\lambda=2$ & $\lambda=20$ & $\lambda=2$ & $\lambda=20$ & $\lambda=2$ &  $\lambda=20$ & $\lambda=20$ & $\lambda=40$ \\
        \midrule
         
         CLPF & 0.444 & -0.698 & -0.831 & -1.939 & 1.322 & -0.077 & -2.620 & -3.963 \\
         CLPF (Particle Filter) & {\bf 0.423} & {\bf -0.756} & {\bf -0.840} & {\bf -1.985} & {\bf 1.213} & {\bf -0.197} & {\bf -2.647} & {\bf -3.966} \\
         \midrule
         Latent SDE & {\bf 1.243} & 1.778 & 0.082 & 0.217 & 3.594 & 3.603 & 7.740 & 8.256 \\
         Latent SDE (Particle Filter) & 1.263 & {\bf 1.001} & {\bf 0.053} & {\bf 0.153} & {\bf 3.573} & {\bf 3.470} & {\bf 7.728} & {\bf 8.255} \\
        \bottomrule
    \end{tabular}}\\
    \label{tab:likelihood}
\end{table}
\begin{table}[t]
    \caption{\small {\bf Sequential Prediction}. We report the average L2-distance between prediction results and ground truth observations in a sequential prediction setting. All predictions are based on the average of 125 latent samples. For evaluations without particle filtering, we sample from the learned variational posterior process.
    }   
    \centering
    \small
    \scalebox{0.85}{
    \begin{tabular}{l@{\extracolsep{3pt}}rrrrrrrr@{}}
        \toprule
         \multirow{2}{*}{Model}
         &\multicolumn{2}{c}{GBM}&\multicolumn{2}{c}{LSDE}&\multicolumn{2}{c}{CAR}&\multicolumn{2}{c}{SLC}\\
         \cmidrule{2-3}
         \cmidrule{4-5}
         \cmidrule{6-7}
         \cmidrule{8-9}
         &$\lambda=2$ & $\lambda=20$ & $\lambda=2$ & $\lambda=20$ & $\lambda=2$ &  $\lambda=20$ & $\lambda=20$ & $\lambda=40$ \\
        \midrule
         
         CLPF & 0.705 & {\bf 0.206} & {\bf 0.102} & {\bf 0.031} & 1.322 & 0.273 & 0.446 & {\bf 0.231} \\
         CLPF (Particle Filter) & {\bf 0.693} & {\bf 0.206} & 0.103 & {\bf 0.031} & {\bf 0.753} & {\bf 0.119} & {\bf 0.422} & 0.236 \\
         \midrule
         Latent SDE & 1.836 & {\bf 1.066} & 0.302 & 0.177 & 63.750 & 57.212 & 14.356 & 14.210 \\
         Latent SDE (Particle Filter) & {\bf 1.503} & 1.282 & {\bf 0.202} & {\bf 0.154} & {\bf 45.403} & {\bf 42.451} & {\bf 14.137} & {\bf 13.663} \\
        \bottomrule
    \end{tabular}
    }
    \label{tab:prediction}
\end{table}

\subsection{Quantitative Results}
Our likelihood evaluation and sequential prediction results are shown in Table~\ref{tab:likelihood} and Table~\ref{tab:prediction}, respectively. We observe that the inference results based on particle filtering are better than the results obtained without using particle filters in almost all settings (81\% of test cases), regardless of model, task, and dataset. Our results in likelihood estimation tasks indicate that particle filtering algorithms generally perform better in complex settings, including GBM with $\lambda=20$ and CAR, which is a non-Markov process. In the sequential prediction tasks we observe a similar trend, especially when the value of $\lambda$ is small and the observations are sparse. In the experiment settings where inference based on particle filtering does not outperform the baselines, the performance of particle filtering-based inference is still competitive. For large values of $\lambda$, previous observations can be close to the observation at the next time point because the change of a continuous process is small during a short time interval, constraining the solution space. We hypothesize that this is also the reason why we see similar prediction accuracies of CLPF models with and without particle filtering with these settings, e.g., for GBM ($\lambda=20$) and LSDE ($\lambda=[2,20]$), as CLPF makes future predictions based on continuous trajectories extrapolated from given observations.

\subsection{Qualitative Study}
To obtain further insights into the quantitative improvements brought about by continuous-time particle filtering, we conduct two qualitative studies on CLPF models using data simulated from the continuous auto-regressive process (CAR), where we observe the most significant improvements.

\begin{figure*}[t]
    \centering
    \begin{subfigure}[b]{0.45\textwidth}
    \includegraphics[width=0.95\textwidth]{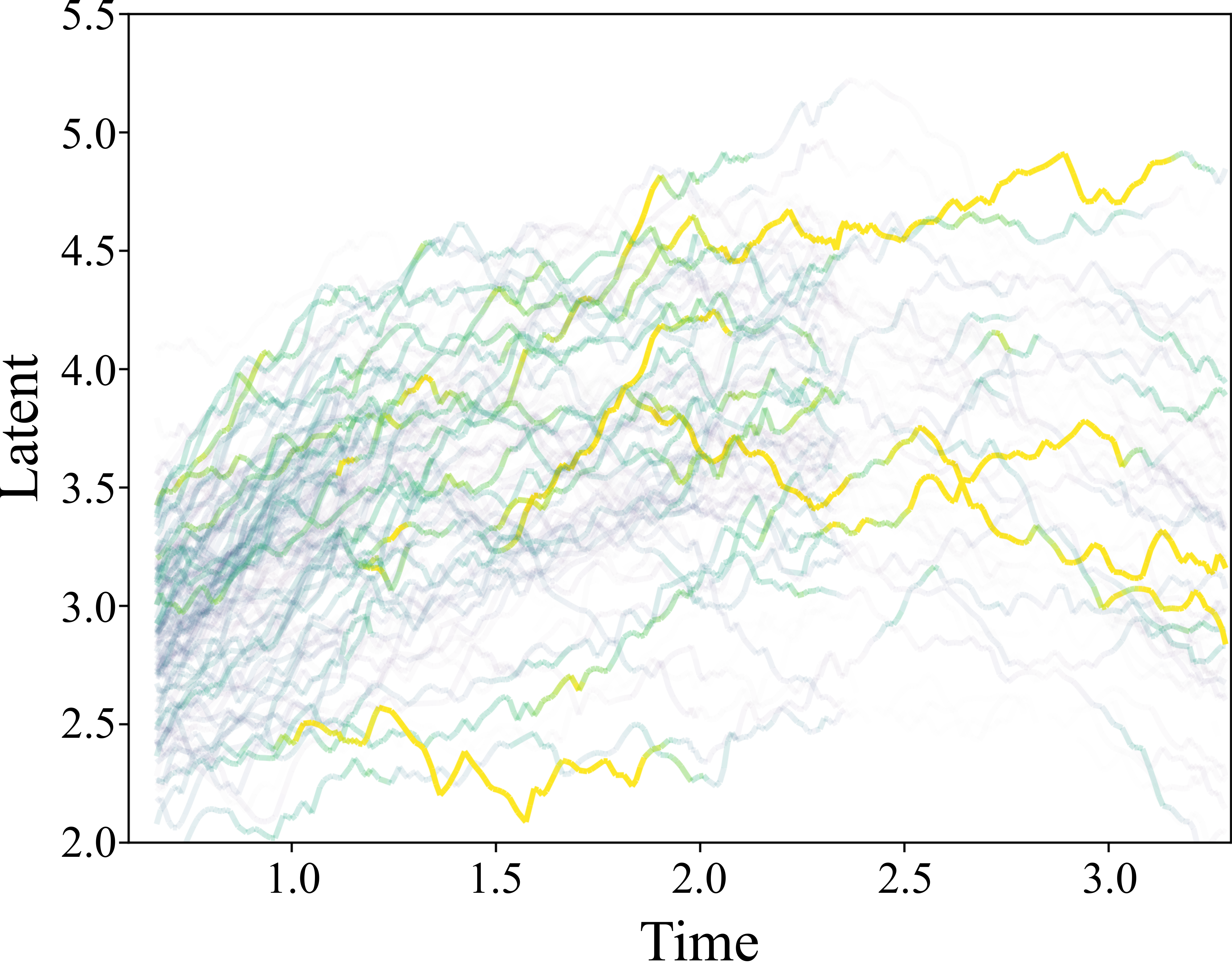}
    \caption{Without Particle Filtering}
    \label{subfig:weight}
    \end{subfigure}
    \begin{subfigure}[b]{0.45\textwidth}
    \includegraphics[width=0.95\textwidth]{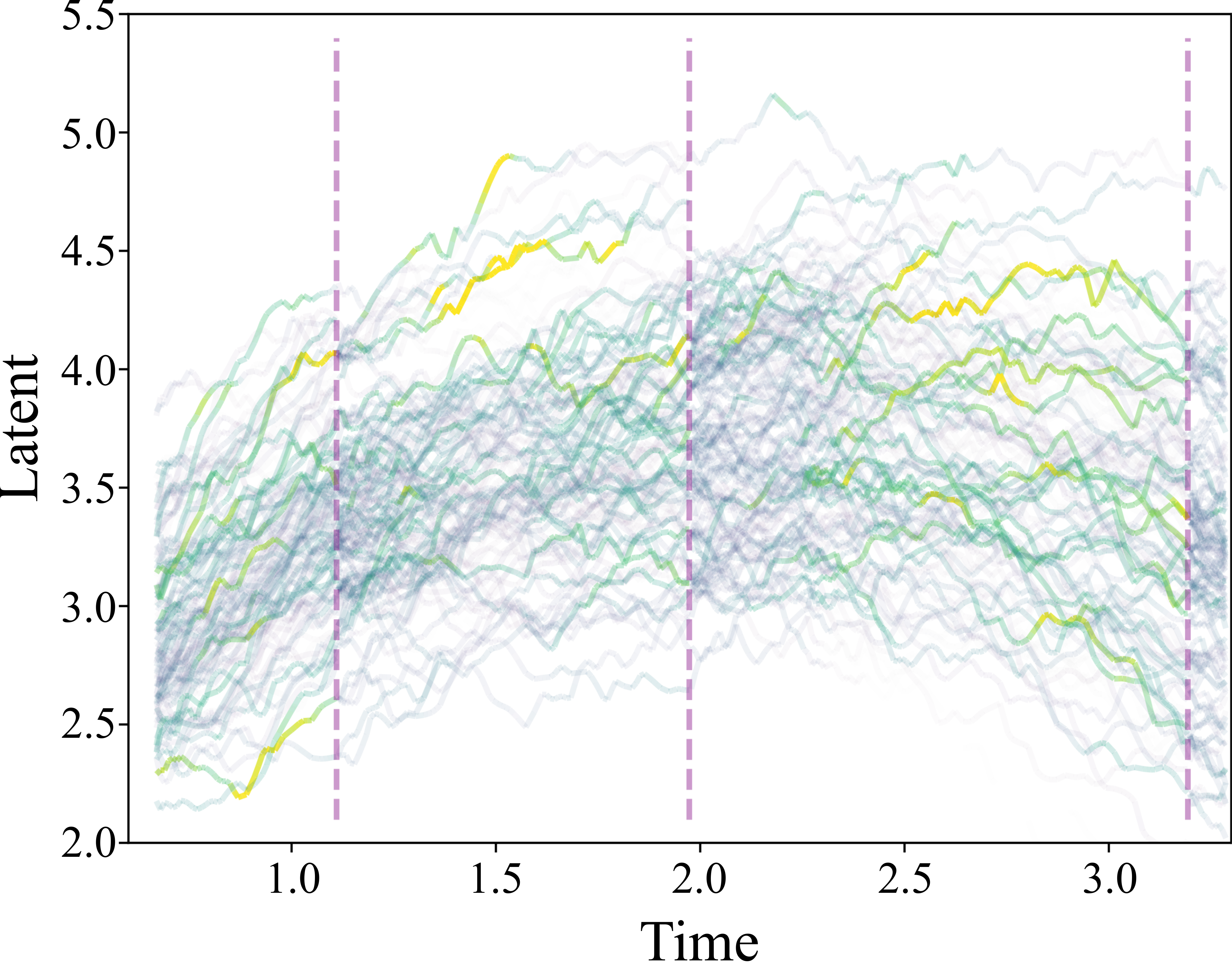}
    \caption{With Particle Filtering}
    \label{subfig:weight_pf}
    \end{subfigure}
    \caption{\small{\bf Qualitative Evaluation (Particle Weights).} We show a comparison between the weights of latent trajectories in a CAR process without (left) and with (right) particle filtering. 
    Transparency and color are used to encode the weights. More transparent segments indicate smaller weights than less transparent segments. Yellow indicates larger weights than green than grey. Dashed purple vertical lines indicate particle resampling. The resampling step of our continuous-time \mbox{particle filtering approach prevents particle decay.}}
    \label{fig:weights}
\end{figure*}

In Figure~\ref{fig:weights}, we visualize the weights of latent trajectories (i.e., samples/particles) over time without (Figure~\ref{subfig:weight}; equivalent to IWAE) and with (Figure~\ref{subfig:weight_pf}; ours) the advantages of particle-based resampling. Both experiments use the exact same sequence of observations as inputs. Grey colors indicate smaller and yellow colors indicate larger weights, demonstrating that particle weights based on particle filtering are less skewed, have lower variance, and do not decay as much over time. We can also observe particles with small weights getting dropped during the resampling step (e.g., at time $t = 2.0$), resulting in discontinued trajectories in Figure~\ref{subfig:weight_pf}. Our comparisons underpin the better sampling efficiency of particle filtering compared \mbox{to an IWAE approximating the likelihood.}

\begin{figure*}[t]
    \centering
    \begin{subfigure}[b]{0.45\textwidth}
    \includegraphics[width=0.95\textwidth]{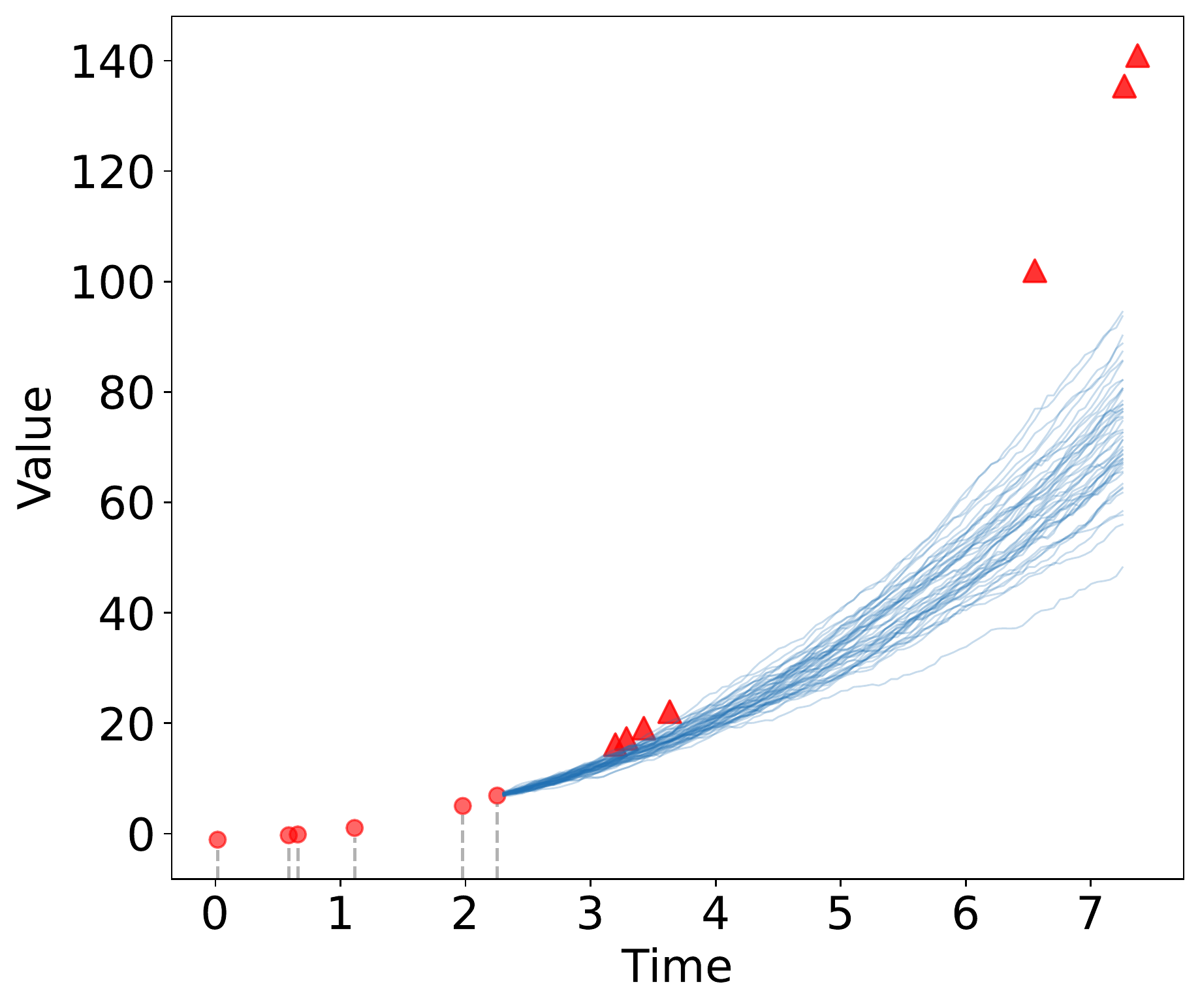}
    \caption{Prediction using Variational Posterior}
    \label{subfig:pred}
    \end{subfigure}
    \begin{subfigure}[b]{0.45\textwidth}
    \includegraphics[width=0.95\textwidth]{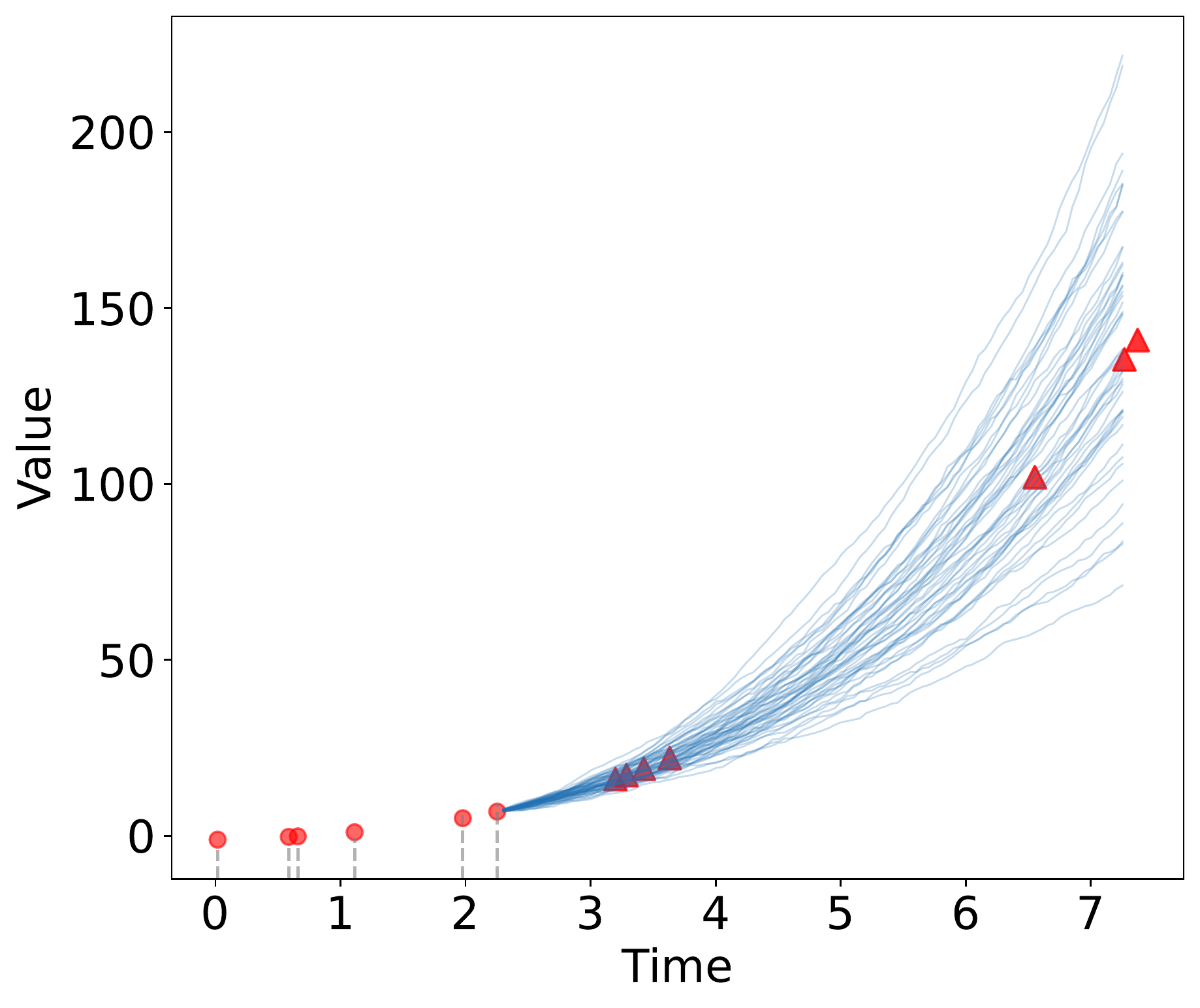}
    \caption{Prediction using Particle Filtering}
    \label{subfig:pred_pf}
    \end{subfigure}
    \caption{\small{\bf Qualitative Evaluation (Approximation Accuracy).} We show extrapolated trajectories (blue) conditioned on a CAR observation sequence (red dots) using a variational (left) and a particle filtering (right) approach. Simulated realizations (red triangles) from the ground truth process \mbox{demonstrate the superior accuracy of particle filtering.}}
    \label{fig:pred}
\end{figure*}

In Figure~\ref{fig:pred}, we compare continuous trajectories extrapolated into the future (blue), conditioned on discrete past observations (red dots). We sample the latent process using two different approaches: (1) by fully relying on the learned variational posterior process, as is usual in latent variable model inference (Figure~\ref{subfig:pred}); and (2) by leveraging the proposed particle filtering approach, i.e., sampling from the posterior process conditioned on the observations and extrapolating the sampled trajectories into the future using the prior process (Figure~\ref{subfig:pred_pf}). We also include data directly simulated from the ground truth process (red triangles) in both figures, confirming that extrapolation based on particle filtering is closer to the true generative process than the learned variational posterior process.

\section{Related Work}\label{sec:related}
As a particle-based inference technique for latent SDEs, our approach is most closely related to previous work in the area of neural differential equations and particle filtering.

\paragraph{Neural Differential Equations.}
The introduction of neural ordinary differential equations (neural ODEs;~\citep{chen2018neural}) has ignited a new area of research and created interesting questions related to the training and inference of neural differential equations. Chronologically, neural architectures leveraging \emph{ordinary} differential equations predate their stochastic siblings: ODE-RNNs~\citep{rubanova2019latent} model the hidden dynamics between consecutive RNN steps with ODEs to reflect non-uniform intervals between observations. A drawback of ODEs is that their solutions depend solely on an initial value, a limitation that has been addressed with neural controlled differential equations (neural CDEs;~\citep{kidger2020neural}) and neural rough differential equations (neural RDEs;~\citep{morrill2021neuralrough}), enabling a flexible adaptation to new data. In addition to data-dependent responses, neural ODE processes (NDPs;~\citep{norcliffe2021neural}) define a distribution over neural ODEs and allow reasoning about the uncertainty associated with latent dynamics. In the context of continuous normalizing flows, \citep{deng2020modeling} propose a differential deformation of a Wiener base process driven by an neural ODE. The case of sporadic observations in multivariate time-series has been addressed with a continuous-time variant of the gated recurrent unit (GRU;~\citep{cho14gru}) and associated Bayesian updates (GRU-ODE-Bayes;~\cite{de2019gru}).

The continuous-time particle filtering approach proposed in this paper is most directly applicable to neural \emph{stochastic} differential equations: \cite{tzen2019neural} view neural SDEs as the diffusion limit of deep latent Gaussian models (DLGMs;~\citep{rezende14dlgm}) and leverage Girsanov reparameterization to derive a mean-field approximation for neural SDEs. Gradient-based optimization in this framework depends on computationally expensive forward simulations of SDEs, a drawback that has been addressed with the scalable gradients of the stochastic adjoint sensitivity method~\cite{li2020scalable}. Continuous Latent Process Flows (CLPFs;~\citep{deng2021continuous}) leverage latent SDEs for continuous-time indexing of a time-dependent flow decoder and introduce a piecewise construction of the variational posterior process. \citep{kidger2020neural_gan} formalize the insight that neural SDEs and Wasserstein GANs both transform noise into data and express traditional SDE training as a special case of a learnt discriminator statistic.
    
\paragraph{Particle Filtering.} Introduced in a seminal work by~\citep{gordon93smc} and a quasi-successor to sequential importance sampling (SIS) and sampling and importance resampling (SIR;~\citep{rubin87SIR}), particle filtering has found applications in numerous inference tasks with intractable state space. Initially viewed as an alternative to extended/unscented Kalman filtering (EKF~\citep{jazwinski70}; UKF~\citep{julier97}) of non-linear/non-Gaussian dynamical systems, it has since been generalized to particle representations of variables~\citep{koller99} or messages~\citep{sudderth03} in more general probabilistic graphical models~\citep{koller09}, and utilized in training neural networks~\cite{freitas00}. A finite-sample analysis using concentration bounds and a derivation of convergence rates for such systems can be found in~\citep{ihler09bp}. Using EKF/UKF approximations as the proposal distributions for particle filters has been explored in~\citep{NIPS2000_f5c3dd75}, as well as extensions for marginalizing out variables in high-dimensional spaces~\cite{doucet2010rao}. Applications of particle filters to continuous-time models are rare and limited to traditional methods without deep architecture or gradient-based optimization: \citep{ng2005continuous} describe particle filtering in a hybrid-state process in which a discrete state variable evolves according to a continuous-time Markov Jump process and show applications to state estimation tasks of a Mars rover; \citep{murray2007continuous} construct a continuous-time filtering framework for hemodynamic interactions in the brain. In the context of variational inference, particle filtering has been used to leverage sequential structure and obtain tighter lower bounds of the marginal likelihood~\cite{maddison2017filtering}.
\section{Discussion and Conclusion}
While continuous-time particle filtering can result in significant performance improvements, there are also limitations that should be addressed in future work. One such limitation is the inference speed of particle filtering, which is slower than techniques without particle filtering. Every time the weights of particles are updated with a new observation, the particle filtering algorithm will also check whether a resampling should be triggered, and resample the particles if necessary. Therefore, continuous-time particle filters lose opportunities for parallelization during certain inference steps, such as decoding, and incur larger computational overhead when solving the SDE. Furthermore, applying particle filtering to the training of deep sequential models has always been a challenging task as gradients cannot be directly backpropagated through the resampling step. Training latent SDE models using continuous-time particle filtering also faces this challenge and would require special techniques, such as Gumbel-softmax or policy gradients.

In this work we extended the popular particle filtering framework from the traditional discrete time series setting to continuous-time latent SDE models. We proposed a mathematically rigorous framework for continuous-time particles, including importance weighting and update schemes, and demonstrated how to leverage the proposed framework in two common inference tasks, likelihood evaluation and sequential prediction, where particle filtering can be used as a drop-in replacement for inference based on the variational posterior. The effectiveness and general applicability of continuous-time particle filtering has been shown on two models in the latent SDE family, CLPF and latent SDE, and four continuous-time stochastic processes.

{\small
  \bibliographystyle{plainnat}
  \bibliography{references}
}
\clearpage
\appendix
\section{Piece-Wise Construction of Posterior Process}
This section provides a complete and detailed description of the piece-wise approach to the construction of the posterior process.
Our piece-wise construction of the posterior process is based on the following fact of Wiener process:   
Given a time grid $0=t_0<t_1<\dots<t_n=T$, we can sample a Wiener process trajectory of length $T$ via sampling from $n$ independent Wiener processes $\{\mW_t^{(i)}\}_{i=1}^n$, each of length $t_i - t_{i-1}$ defined on the filtered probability space $(\Omega^{(i)}, \mathcal{F}^{(i)}_t, P^{(i)})$ with distribution $P^{(i)}$, in the following way:
\begin{equation}
    \mW_t(\omega^{(1)}, \omega^{(2)}, \dots, \omega^{(n)}) = \sum_{\{i: t_i<t\}}\mW_{t_i- t_{i-1}}^{(i)}(\omega^{(i)}) + \mW_{t- t_{i^*}}^{(i^*)}(\omega^{(i^*)})
\end{equation} where $i^* = \max\{i: t_i<t\} + 1$ and $\omega^{(i)}\in \Omega^{(i)}$.
This construction of Wiener process allows us to solve the stochastic differential equation
\begin{equation}
    \textrm{d}\mZ_t = \mu_\theta(\mZ_t, t)\ \textrm{d}t + \sigma_\theta(\mZ_t, t)\ \textrm{d}\mW_t
\end{equation} and sample $\{\mZ_{t_i}\}_{i=1}^n$ in a piece-wise manner given $\mZ_{t_0}$.
Taking $\mZ_{t_{i-1}}$ as the initial condition, we can sample $\mZ_{t_i}$ by solving the SDE
\begin{equation}
    \textrm{d}\mZ_t = \mu_\theta(\mZ_t, t)\ \textrm{d}t + \sigma_\theta(\mZ_t, t)\ \textrm{d}\mW_t^{(i)}
\end{equation}
in the interval between $t_{i-1}$ and $t_i$, i.e.
\begin{equation}
    \mZ_{t_i} =  \mZ_{t_{i-1}}+ \int_{t_{i-1}}^{t_i} \mu_\theta(\mZ_s,s )\ \textrm{d}s + \int_{t_{i-1}}^{t_i} \sigma_\theta(\mZ_s,s )\ \textrm{d}\mW^{(i)}_{s - t_{i-1}}.
\end{equation}
Therefore we rewrite the expectation on the left-hand side of Eq. 7 in Section 3.1 as 
\begin{equation}
    \mathbb{E}_{P^{(1)}}\left[\dots\mathbb{E}_{P^{(i)}}\left[\dots \mathbb{E}_{P^{(n)}}\left[ f\left(\left\{\mZ_{t_i}\right\}_{i=1}^n\right)\right]\dots\right]\dots\right].
    \label{eq:exp_nest}
\end{equation}
For the simplicity of presentation, we rewrite the term in the expectation  $\mathbb{E}_{P^{(i)}}\left[\cdot\right]$ as a function of $\mZ_{t_i}$ conditioned on $\left\{\mZ_{t_k}\right\}_{k=1}^{i-1}$, i.e.,
\begin{equation}
    \mathbb{E}_{P^{(i)}}\left[\dots \mathbb{E}_{P^{(n)}}\left[ f\left(\left\{\mZ_{t_i}\right\}_{i=1}^n\right)\right]\dots\right]=\mathbb{E}_{P^{(i)}}\left[f^{(i)}(\mZ_{t_i}|\left\{\mZ_{t_j}\right\}_{j=1}^{i-1}) \right]
\end{equation}
for some function $f^{(i)}(\cdot|\left\{\mZ_{t_j}\right\}_{j=1}^{i-1})$ thanks to the Markov property of SDE solutions.
For each interval $[t_{i-1}, t_i]$, we can define a posterior process SDE 
\begin{equation}
    \textrm{d}\tilde{\mZ_t} = \mu_{\phi^{(i)}}(\tilde{\mZ_t}, t)\ \textrm{d}t + \sigma_\theta(\tilde{\mZ_t}, t)\ \textrm{d}\mW_t^{(i)}
\end{equation} with $\phi^{(i)}$ potentially being (partially) parameterized by $\left\{\mZ_{t_k}\right\}_{k=1}^{i-1}$ and define distribution $Q^{(i)}$ for $(\Omega^{(i)}, \mathcal{F}^{(i)}_t)$ according to Girsanov Theorem~\cite{oksendal2013stochastic} such that
\begin{align}
    \begin{split}
    &\mathbb{E}_{P^{(i)}}\left[f^{(i)}(\mZ_{t_i}|\left\{\mZ_{t_k}\right\}_{k=1}^{i-1}) \right] \\[4pt]
    =& \mathbb{E}_{Q^{(i)}|\left\{\mZ_{t_k}\right\}_{k=1}^{i-1}}\left[f^{(i)}(\tilde{\mZ}_{t_i}|\left\{\mZ_{t_k}\right\}_{k=1}^{i-1})\right]\\[4pt]
    =&\mathbb{E}_{P^{(i)}}\left[f^{(i)}(\tilde{\mZ}_{t_i}|\left\{\mZ_{t_k}\right\}_{k=1}^{i-1})\mM^{(i)} \right]
    \end{split}
\end{align}
Please refer to Section 3.1 for the details of defining $Q^{(i)}$ and $\mM^{(i)}$. As a result, Eq.~\ref{eq:exp_nest} can be rewritten in the following way
\begin{align}
\begin{split}
    &\mathbb{E}_{P^{(1)}}\left[\dots\mathbb{E}_{P^{(i)}}\left[\dots \mathbb{E}_{P^{(n)}}\left[f\left(\left\{\mZ_{t_i}\right\}_{i=1}^n\right)\right]\dots\right]\dots\right] \\[4pt]
    =\ &\mathbb{E}_{Q^{(1)}|\{\tilde{\mZ}_{t_1}\}}\left[\dots\mathbb{E}_{Q^{(i)}|\{\tilde\mZ_{t_k}\}_{k=1}^{i-1}}\left[\dots \mathbb{E}_{Q^{(n)}|\{\tilde\mZ_{t_k}\}_{k=1}^{n-1}}\left[f(\{\tilde\mZ_{t_i}\}_{i=1}^n)\right]\dots\right]\dots\right]\\[4pt]
    =\ &\mathbb{E}_{P^{(1)}}\left[\dots\mathbb{E}_{P^{(i)}}\left[\dots \mathbb{E}_{P^{(n)}}\left[f(\{\tilde\mZ_{t_i}\}_{i=1}^n)\mM^{(n)}\right]\dots \mM^{(i)}\right]\dots\mM^{(1)}\right].
\end{split}
\label{eq:exp_piece_supp}
\end{align}

For latent SDE model~\cite{li2020scalable}, $\phi^{(i)}$ is completely determined by observation sequence $\{\vx_{t_i}\}_{i=1}^n$ and remains the same for each time interval $[t_{i-1}, t_i]$. In CLPF~\cite{deng2021continuous} models, $\phi^{(i)}$ is parameterized by $\{\vx_{t_k}\}_{k=1}^i$ and $\{\mZ_{t_k}\}_{k=1}^{i-1}$ in the interval between $t_{i-1}$ and $t_i$

\section{Parameters of Stochastic Processes for Data Simulation}
In our experiments, data sequences are sampled from four common continuous stochastic processes: geometric Brownian motion (GBM), linear SDE (LSDE), continuous auto-regressive process (CAR), and stochastic Lorenz curve (SLC). The sequences are simulated using the Euler-Maruyama method~\cite{bayram2018numerical} with a fixed step size of $1e-5$ in the time interval $[0, 30]$ for GBM, LSDE, and CAR and in the time interval $[0,2]$ for SLC. 
Below are the parameters we use for simulation in each process:

\textbf{Geometric Brownian Motion.} Observations are sampled from the SDE $\textrm{d}\mX_t = 0.2\mX_t\ \textrm{d}t + 0.1\mX_t\ \textrm{d}\mW_t$, with an initial value $\mX_0=1$.

\textbf{Linear SDE.}
We simulate sequences from the the SDE $\textrm{d}\mX_t = (0.5\sin(t)\mX_t + 0.5\cos(t))\ \textrm{d}t + \frac{0.2}{1+\exp(-t)}\ \textrm{d}\mW_t$ with initial value 0.
 
\textbf{Continuous AR(4) Process.} A CAR process $\mX_t$ can be viewed as the linear projection of a process defined by a high-dimensional SDE to a low-dimensional one. The high-dimensional SDE and linear projection in our fourth-order CAR process are
   \begin{align}
   \small
        \begin{split}
        \begin{split}
             \textrm{d}\mY_t &= A\mY_t\ \textrm{d}t + e \ \textrm{d}\mW_t,\\
             \mX_t &= [1, 0, 0, 0]\mY_t, 
        \end{split} \quad \mbox{ where } 
        \tiny
        A = \left[\begin{matrix}0 & 1 &0&0\\
        0&0&1&0\\
        0&0&0&1\\
        0.002&0.005&-0.003&-0.002
        \end{matrix}\right] \text{\small and }
    e = [0, 0, 0, 1].
    \end{split}
    \end{align}
\textbf{Stochastic Lorenz Curve.} The stochastic Lorenz curve is defined by the following three-dimensional SDE
\begin{align}
        \begin{split}
            \textrm{d}\mX_t &= 10(\mY_t - \mX_t)\ \textrm{d}t + 0.1\ \textrm{d}\mW_t, \\
            \textrm{d}\mY_t &= (\mX_t(28 - \mZ_t) - \mY_t)\ \textrm{d}t + 0.28\ \textrm{d}\mW_t, \\
            \textrm{d}\mZ_t &= (\mX_t\mY_t - \frac{8}{3}\mZ_t)\ \textrm{d}t + 0.3\ \textrm{d}\mW_t. \\
        \end{split}
    \end{align}

\section{Additional Experiment Setting Details}
\label{sec:additional_experiment_details}
The experiment settings and model architectures are aligned with the synthetic data evaluation settings of CLPF~\cite{deng2021continuous}. The CLPF and latent SDE models are pretrained using 7000 sequences for training and 1000 sequence for validation. The training and validation data are simulated with $\lambda=2$ for GBM, LSDE, and CAR and $\lambda=20$ for SLC. We use IWAE bound estimated by 3 latent samples during training and 125 latent samples (particles) during evaluation.

The sequential prediction task can be characterized as predicting $\vx_{t_i}$ given observations $\{\vx_{t_k}\}_{k=1}^{i-1}$. When using latent SDE without particle filtering for sequential prediction, we directly sample $\mZ_{t_i}$ from the posterior process 
\begin{equation}
    \textrm{d}\mZ_t = \mu_\phi(\mZ_t, t)\ \textrm{d}t + \sigma_\theta(\mZ_t, t)\ \textrm{d}\mW_t
\end{equation}
with $\phi$ parameterized by observation $\{\vx_{t_k}\}_{k=1}^{i-1}$ and take the average over sample of $\mX_{t_i}$ decoded from $\mZ_{t_i}$. When using the plain CLPF models for sequential prediction without particle filtering, we sample $\mZ_{t_i}$ from the posterior process
\begin{equation}
    \textrm{d}\mZ_t = \mu_\phi^{(i-1)}(\mZ_t, t)\ \textrm{d}t + \sigma_\theta(\mZ_t, t)\ \textrm{d}\mW_t
\end{equation}
with $\phi^{(i-1)}$ parameterized by $\{\vx_{t_k}\}_{k=1}^{i-1}$ and samples of $\{\mZ_{t_k}\}_{k=1}^{i-2}$ and also take the average over the decoded samples of $\mX_{t_i}$ as the prediction result. When using particle filtering in the sequential prediction task, we can obtain samples of $\mZ_{t_{i-1}}$ using particle filtering and sample from $\mZ_{t_i}$ using the prior process with $\mZ_{t_{i-1}}$ as the initial condition. The prediction is obtained by taking a weighed average of the decoded $\mX_{t_i}$ samples with weights inherited from their underlying particles.
\end{document}